\documentclass{article}

\usepackage{microtype}
\usepackage{graphicx}
\usepackage{subfigure}
\usepackage{booktabs} 
\usepackage{caption}
\usepackage{wrapfig}
\usepackage{algorithm}
\usepackage{algorithmic}

\usepackage{amsmath,amsfonts,bm}

\def\eqref#1{equation~\ref{#1}}

\def\1{\bm{1}}

\DeclareMathAlphabet{\mathsfit}{\encodingdefault}{\sfdefault}{m}{sl}
\SetMathAlphabet{\mathsfit}{bold}{\encodingdefault}{\sfdefault}{bx}{n}

\newcommand{\R}{\mathbb{R}}

\newcommand{\softmax}[1]{\mathrm{Softmax}\left(#1\right)}
\newcommand{\lstm}[1]{\mathrm{LSTM}\left(#1\right)}
\newcommand{\attention}[1]{\mathrm{Attention}\left(#1\right)}
\newcommand{\concat}[1]{\mathrm{Concat}\left(#1\right)}

\newcommand{\norm}[2]{||#1||_{#2}}

\def\programs{{\mathcal{Y}}}
\def\specs{{\mathcal{X}}}

\newcommand{\aef}{\mathrm{ec}}
\newcommand{\lp}{\mathrm{lp}}
\newcommand{\vqf}{\mathrm{qc}}
\newcommand{\vqk}{\mathrm{qk}}

\usepackage{url}
\usepackage{multirow}
\usepackage{makecell}
\usepackage[T1]{fontenc}

\title{Latent Programmer: Discrete Latent Codes for Program Synthesis}

\author{
Joey Hong \And David Dohan \And  Rishabh Singh \And Charles Sutton \And Manzil Zaheer
}

\newcommand{\logicalOR}{\; | \;}
\newcommand{\T}[1]{\texttt{#1}}

\newcommand{\var}[1]{\textsf{#1}}
\newcommand{\calT}[0]{\mathcal{T}}

\usepackage{hyperref}

\usepackage[accepted]{icml2021}

\icmltitlerunning{Latent Programmer: Discrete Latent Codes for Program Synthesis}

\begin{document}

\twocolumn[
\icmltitle{Latent Programmer: Discrete Latent Codes for Program Synthesis}



\icmlsetsymbol{equal}{*}

\begin{icmlauthorlist}
\icmlauthor{Joey Hong}{goo}
\icmlauthor{David Dohan}{goo}
\icmlauthor{Rishabh Singh}{goo}
\icmlauthor{Charles Sutton}{goo}
\icmlauthor{Manzil Zaheer}{goo}
\end{icmlauthorlist}

\icmlaffiliation{goo}{Google Research, Mountain View, CA, USA}

\icmlcorrespondingauthor{Joey Hong}{jxihong@google.com}

\icmlkeywords{Machine Learning, ICML}

\vskip 0.3in
]



\printAffiliationsAndNotice{\icmlEqualContribution} 

\begin{abstract}
A key problem in program synthesis is searching over the large space of possible programs.
Human programmers might decide the high-level structure of the desired program before thinking
about the details;
motivated by this intuition, we consider two-level search for program synthesis, in which the synthesizer
first generates a plan---a sequence of symbols that describes the desired program at a high level---before generating the program.
We propose to learn representations of programs that can act as plans to organize such a two-level search. 
Discrete latent codes are appealing for this purpose, and can be learned by applying recent work on discrete autoencoders.
Based on these insights, we introduce the \emph{Latent Programmer} (LP), a program synthesis method that first predicts a discrete latent code from input/output examples, and then generates the program in the target language. 
We evaluate the LP on two domains,
demonstrating that it yields an improvement in accuracy, especially on longer programs for which search is most difficult.
\end{abstract}

\vspace{-2mm}
\section{Introduction}
\vspace{-1mm}



Program synthesis is a longstanding grand challenge in artificial intelligence~\citep{mannaw71,summers1977methodology}.
The objective of program synthesis is to automatically write a program given
a specification of its intended behavior, such as a natural language description or a small set of  input-output examples~\citep{sygus,now17}.
However, program synthesis requires solving a difficult search problem
over a large space of possible programs.
Search methods
that have been explored include top-down search \citep{EUPHONY}, bottom up search \citep{TRANSIT,BUSTLE,Barke2020-rq},
 beam search  \citep{ROBUSTFILL},
and many others (see~\autoref{sec:background}).

Our work is motivated by an intuition about the way people write
programs.
Synthesis methods often search through programs
in an order determined by the token sequence,
by a syntax tree, or by a logical solver.
In contrast, we imagine that a programmer often starts by thinking about the high-level structure of the desired program --- such as what library functions
to call, or the overall program structure --- and then fills in the details.
For example, for a program that processes a list of people's names, 
a programmer might first plan that the program should extract the person's name followed by the person's last initial, and only then think about details such as  which library functions to use.

Based on this intuition, we propose \emph{two-level search} for program synthesis.
In two-level search, the synthesizer first produces a 
plan that describes the desired program, and then synthesizes
a program based on that plan.
For us, a \emph{plan} is simply a sequence of symbols that describes the code to be generated at a high level, 
without specifying the syntactic and semantic details.
The goal is that plans can provide a way to organize search over programs.
In the name-processing example, suppose that an initial plan  incorrectly specifies to extract the first instead of the last initial.
Searching in plan space could easily find the small change required to yield the correct plan, even if this would correspond to a large change in the program. This allows two-level search to explore a more diverse set of programs, improving the chance of finding the correct one.

A key design decision is defining the space of possible plans.
For example, sketches could be used as plans \cite{SKETCHADAPT, Bayou}.
In this work, we explore whether it is possible to learn a representation
of programs that is useful for constructing plans that guide search.
Instead of having a deterministic heuristic for mapping programs to plans, 
we let a model discover what plans are useful for representing programs,
and how to infer them from the specification.
To tackle this problem, we make use of
recent work in learning discrete unsupervised representations
\citep{VQVAE,Roy2018,Kaiser2018}. These are self-supervised methods
that given a dataset, assign each data item to a \emph{discrete latent code} (a sequence of symbols from an arbitrary set) in such a way 
that the latent code provides a good description of the data item.
The main hypothesis of our work is that discrete latent codes
can be used as plans for two-level search.



This leads us to propose the \emph{Latent Programmer},
a program synthesis method that employs
two-level beam search, where the plans are based on discrete latent codes.
At training time, a discrete autoencoder based on \citet{Kaiser2018}
is used to train three models: one that infers ground-truth discrete latent codes to describe the programs in the training set, one that
maps specifications to plans (i.e. discrete latent codes), and one that maps plans to programs.
At inference time, Latent Programmer uses a two level beam search,
first 
producing an $L$-best list of plans, then producing a $B/L$-best list of programs
for each plan.
On two different program synthesis domains,
we find empirically that the Latent Programmer
improves synthesis accuracy by over $10\%$ compared to several baseline synthesis methods, 
especially on longer programs 
that are more difficult for search.


\section{Background}
\label{sec:background}
\vspace{-1mm}

\begin{figure*}[ht]
\centering
\begin{tabular}{l l l}
 \Xhline{2\arrayrulewidth}
 Inputs & Outputs & Program\\[0.1cm]
 \hline
 ``Mason Smith" &  ``Smith M" &  \\
 ``Henry Myers" & ``Myers H" & \small\texttt{GetToken\_PROP\_CASE\_2 | Const(`` ") |}\\
 ``Barry Underwood" & ``Underwood B" & \small\texttt{GetToken\_ALL\_CAPS\_1} \\
 ``Sandy Jones" & ``Jones S" \\
\Xhline{2\arrayrulewidth}
\end{tabular}
\caption{A string transformation task with four input-output examples and a program in the DSL that is consistent with the examples.}
\label{fig:robustfill_example}
\end{figure*}

The goal in program synthesis is to find a program in a given language that is consistent with a specification.
Formally, we are given a domain specific language (DSL) which defines a space $\programs$ of programs. 
The task is described by a specification $X \in \specs$ and is solved by an unknown program $Y \in \programs$. 
For example, each specification can be a set of input/output (I/O) examples denoted $X = \{(I_1, O_1), \hdots (I_N, O_N)\}$. Then, we have solved specification $X$ if we found a program $Y'$ (not necessarily $Y$) which correctly solves all the examples: $Y'(I_i) = O_i, \, \forall i$. As another example, each specification can be a natural language description of a task, and the corresponding program implements said task. An example synthesis task in the string transformation DSL is shown in \autoref{fig:robustfill_example}.

\textbf{Vector Quantization}
\label{sec:background_vq}
Our method relies on \emph{discrete autoencoders}, which are unsupervised learning methods that
assign each data point to a sequence of symbols, called a \emph{discrete latent code}, in such
a way that the code is a good description of the data item.
In this section, we describe Vector Quantized Variational Autoencoders (VQ-VAE)
\citep{VQVAE,Roy2018}. This will introduce
ideas that we use later in the discrete autoencoder for Latent Programmer (Section~\ref{sec:training}).
In a VQ-VAE, latent codes are sequences 
drawn from a discrete set of tokens $\calT$ of size $|\calT| = K$. 
Each token with id $k \in [K]$ is associated with a learned embedding $c_k \in \mathbb{R}^D;$ these
embeddings can be stacked into a matrix $c \in \mathbb{R}^{K \times D}$ called a \emph{codebook}.  To generate a code for a data item $x$,
first the data point is passed through a neural network $\aef_\phi(x)$ called an encoder,
and the encoder output $e$
is quantized via nearest-neighbor lookup into the codebook. Formally, the quantized token id $\vqk(e)$ and embedding $\vqf(e)$ are 
\begin{align}
\vqf(e) = c_{\vqk(e)} \text{ where } \vqk(e) = \arg\min_{k \in [K]} \norm{e - c_k}{2}.
\label{eq:vq}
\end{align}
For input $x$, the training loss for a VQ-VAE has three terms: a reconstruction loss, a codebook loss that encourages codebook embeddings to be close to
their associated encoded inputs $\aef(x)$, and a commitment loss that encourages the encoded input $\aef(x)$ to ``commit'' to codes i.e. to be close to the discrete code it is quantized to.
In total, the loss is  
\begin{align}
\mathcal{L}(c, \theta, \phi) &= \log p_{\theta}\left(x \mid \vqf(\aef_\phi(x))\right) + \norm{\mathrm{sg}(\aef_\phi(x)) - c}{2}^2 \nonumber \\
&\quad+ \beta \norm{\mathrm{sg}(c) - \aef_\phi(x)}{2}^2, \label{eq:vq_loss}
\end{align}
where $\theta, \phi$ are the parameters of the decoder and encoder, respectively, $\mathrm{sg}(\cdot)$ is the stop gradient operator that fixes the operand from being updated by gradients, and $\beta$ controls the strength of the commitment loss.
To stabilize training, \citet{VQVAE} also proposed removing the codebook loss and set the codebook to an exponential moving average (EMA) of encoded inputs.

\section{Synthesis with Discrete Latent Variables}
\vspace{-1mm}

Latent Programmer is an instance of a general framework for two-level search in program synthesis  (\autoref{sec:two-level-search}). After presenting the general framework, we describe the specific architecture (\autoref{sec:architecture}), training objective (\autoref{sec:training}),
and search method (\autoref{sec:beam-search}) used in Latent Programmer. 

\vspace{-0.05in}
\subsection{Two-level Search}
\label{sec:two-level-search}
\vspace{-0.05in}

\begin{algorithm}[tb]
    \caption{Program synthesis using two-level search}
    \label{alg:two-level-synthesis}
    \hspace*{\algorithmicindent} \textbf{Input:} Specification $X$, search functions $S_0, S_1,$ \\
    \hspace*{4\algorithmicindent} objective functions $f,g$ 
    \begin{algorithmic}[1]
    \STATE $\var{plan} \gets S_0(X, f)$ 
    \STATE $Y' \gets S_1(\var{plan}, X, g)$ 
    \RETURN $Y'$
    \end{algorithmic}
\end{algorithm}

Our approach is based on the generic framework for
defining program synthesizers using two-level search in Algorithm~\ref{alg:two-level-synthesis}. 
This framework is agnostic to the search algorithm and DSL used.
The idea is that the algorithm generates a \emph{plan}, which intuitively, provides a high-level, coarse-grained description of a program to organize the search procedure. For example,
in string editing, a token in a plan might indicate that the program should extract the first numeric substring.
Formally, a plan is simply a sequence of tokens, each drawn from a finite set $\calT$ of size $|\calT| = K$. 
We denote tokens in $\calT$ as \texttt{TOK\_1}, \texttt{TOK\_2}, $\hdots$, \texttt{TOK\_K}.

To define a concrete synthesizer in this framework, we need to specify concrete choices for the
search algorithms $S_0$ and $S_1$ and objective functions $f$ and $g$ that should be used.
The first-level search function $S_0$ returns $\var{plan} \in \calT^S$ that approximately maximizes
the objective $f(\var{plan}, X) \in \R.$
Then, the second-level search function $S_1$ returns a program $Y' \in \programs$ using $\var{plan}$ that maximizes
$g(Y', \var{plan}, X) \in \R.$





Several previous synthesis methods can be seen as examples of this general framework.
For example, \textsc{SketchAdapt} \cite{SKETCHADAPT} can be viewed as an instantiation of this framework where the plans are program sketches,
that is, partial programs in which certain subtrees are replaced by a special \texttt{HOLE} token. Also,
\textsc{Bayou} \cite{Bayou} can be viewed as another instantiation where the plan is a sketch that abstracts expressions
and function calls by their types.
Our Latent Programmer approach is a new instantiation of this framework, described next.

\vspace{-0.05in}
\subsection{Architecture}
\label{sec:architecture}
\vspace{-0.05in}
\begin{figure*}[t]
\centering
\includegraphics[width=0.65\linewidth]{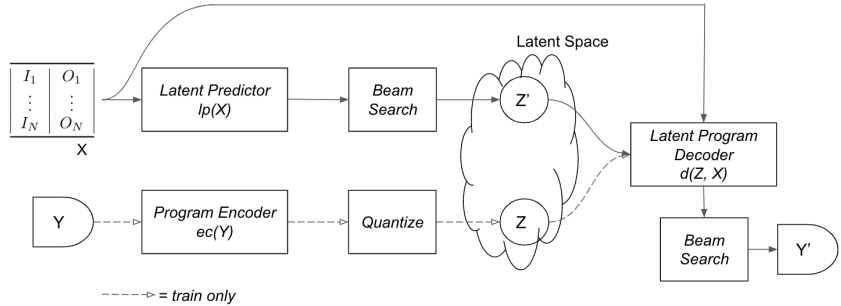}
\caption{High-level architecture for the Latent Programmer system. The latent predictor generates probabilities over latent sequences, which can be decoded into a predicted latent sequence $Z'$. $Z'$ is fitted to a ground-truth latent sequence $Z$ generated by a program encoder, and used during decoding to by the latent program decoder to generate programs.}
\label{fig:lp_architecture}
\end{figure*}

Our synthesizer, \emph{Latent Programmer} (LP), is a two-level synthesizer that learns representations of plans
using a discrete autoencoder.
Because programs are often modular, with components that are reused across tasks,
LP is based on the hypothesis that this compositional structure can be leveraged by learning plans as discrete latent codes.
We use neural networks to define distributions over both plans and programs, which are then used within Algorithm~\ref{alg:two-level-synthesis} through having $f$ and $g$ be the log-probabilities defined by those networks.
In this section, we describe the architecture of LP at a high-level, deferring details  to \autoref{sec:architecture_details}.

Our proposed system consists of three main components: a \emph{latent predictor}, \emph{latent program decoder}, and \emph{program encoder}. Components are parameterized as Transformers, which we use instead of RNNs due to their impressive performance on natural language tasks \citep{Transformer}. 

The pipeline of our LP model is summarized in \autoref{fig:lp_architecture}, and 
an end-to-end example is shown in \autoref{fig:robustfill_example_2}.
The \emph{latent predictor} $\lp(X)$ predicts a distribution over latent codes $\lp(X) \in \mathbb{R}^{S \times K}$ conditioned on the program specification $X$.
The \emph{latent program decoder} $d(Z, X)$ defines a distribution over programs, and is jointly conditioned on specification $X$ and latent code $Z \in \mathbb{R}^{S \times K}$.
The \emph{program encoder} is only used during training and learns useful meanings for the latent tokens in the code.
The program encoder $\aef(Y)$ encodes the true program $Y = [y_1, y_2, \hdots, y_T]$ into a discrete latent code $Z = [z_1, z_2, \hdots, z_S],$ where each $z_i \in \mathcal{T}$. This latent code will then serve as the ground-truth plan for $Y$, as described in the next section.
In this work we let $S = \lceil T / 2^\ell \rceil$, where $\ell$ is the \emph{latent length compression factor} and is tuned during training. This provides temporal abstraction, where the high-level latent tokens roughly map to $2^\ell$ program tokens.
We emphasize that the program encoder is only used in training.  At test time, $\lp(X)$ is used instead of $\aef(Y)$; the latent predictor is unaware of what $S$ is and autoregressively generates latent tokens until an end-of-sequence token is reached.


\vspace{-0.05in}
\subsection{Training}
\label{sec:training}
\vspace{-0.05in}

To learn the plan representations, we use
discrete latent codes from an autoencoder, based on the work of \citet{Kaiser2018} in natural language,
which combines a VQ-VAE with a sequence-to-sequence
learning objective. The loss function
has three parts.

First, the \emph{autoencoder loss} ensures that the latent codes contain information about the program, and that the latent program decoder can  recover the true program given the specification and 
the latent code.
 This loss is similar to the loss function of a VQ-VAE as in \eqref{eq:vq_loss}, but also depends on a specification $X$.
Like in \citet{Roy2018}, the codebook is not trained but set to the EMA of encoder outputs. 
Second, the \emph{latent prediction loss} ensures that latent codes can be predicted from specifications. This loss treats the discrete latent sequence $\vqk(\aef(Y))$ of the true program as the ground-truth plan, and trains the latent predictor $\lp(X)$ to generate it using just the program specification $X$.
Finally, the \emph{end-to-end loss} ensures that programs can be predicted from specifications.
This is needed because when computing the autoencoder loss, the latent code arises from encoding the correct program $\aef(Y)$,
but at test time, we have only the specification $X$.
This can result in mistakes in the generated program since the decoder has never been exposed to noisy results from the latent predictor.
The end-to-end loss alleviates this issue.
To make this differentiable,
the end-to-end loss is probability of the correct program $Y$ when predicted from a soft-quantized latent code, given by
$\lp(X)^T c$.
In summary, the full loss for a training instance is
\begin{align}
&\mathcal{L}(c,\theta, \phi, \psi) \label{eq:lp_loss}  \\
&= \underbrace{\log p_\theta\left(Y \mid \vqf(\aef_\phi(Y)), X \right) + \beta \norm{\mathrm{sg}(c) - \aef_\phi(Y)}{2}^2}_{\text{autoencoder}} \nonumber \\ 
&+ \underbrace{\log p\left(\vqk(\aef_\phi(Y)) \mid \lp_\psi(X)\right)}_{\text{latent prediction}} 
+ \underbrace{\log p_\theta\left(Y \mid \lp_\psi(X)^Tc, X \right)}_{\text{end-to-end}} \nonumber
\end{align}
where we explicitly list out $\theta,\phi,$ and $\psi$ representing the parameters of the latent program decoder, program encoder, and  latent program decoder respectively.

Finally, for the first 10K steps of training, we give embeddings of the ground-truth program $Y$, averaged over every $2^\ell$ tokens,
as the latent code instead of $\aef(Y)$. This pre-training ensures that when we start training
on the full objective, the latent code already contains information about the program that can aid in training the latent program decoder. We found empirically that this prevented the bypassing phenomenon where the latent code is ignored by the decoder \citep{Bypassing}. 

\vspace{-0.1in}
\subsection{Two Level Beam Search}
\label{sec:beam-search}
\vspace{-0.05in}

During inference, we use two-level beam search, i.e., in Algorithm~\ref{alg:two-level-synthesis}, both $S_0, S_1$ are beam search, $f$ is the log probability from the latent predictor, and $g$ the log probability from the latent program decoder.
Standard beam search returns the top-$B$ most likely programs according to the model, from which we return the first one (if 
any) that is consistent with the specification \citep{NeuralProgramSynthesis,ROBUSTFILL}. In our case, $S_0$ performs 
beam search to return $L$ sequences of discrete latent codes, then $S_1$ returns $\lfloor B / L \rfloor$ programs for each latent sequence. During inference, the latent predictor will continue to generate latent tokens until an end-of-sequence token is produced, so  the generated latent sequence does not necessarily  have length $\lceil T / 2^\ell \rceil$ as during training; however, we found the latent sequence lengths during training and evaluation to be close in practice. Setting $L=B$ allows for the maximum exploration of the latent space, while setting $L=1$ reduces our method to standard beam search, or exploitation of the most likely latent decoding. We choose $L=\sqrt{B}$ in our experiments, but explore the effect of $L$ in ~\autoref{sec:analysis}.

\section{Related Work}

\textbf{Program Synthesis } Our work deals with \emph{program synthesis}, which involves combinatorial search for programs that match a specification. Many different search methods have been explored within program synthesis,
including search within a version-space algebra \citep{FLASHFILL},
bottom-up enumerative search \citep{TRANSIT}, stochastic search~\citep{stoke}, genetic programming~\citep{koza1994genetic}, or reducing the synthesis problem to logical satisfiability
\citep{SKETCH}. \emph{Neural program synthesis} involves learning neural networks to predict function distributions to guide a synthesizer \citep{DEEPCODER}, or the program autoregressively in an end-to-end fashion \citep{NeuralProgramSynthesis,ROBUSTFILL}. \textsc{SketchAdapt} \citep{SKETCHADAPT} combined these approaches by first generating a program sketch with holes, and then filling holes using a conventional synthesizer. \textsc{Bayou} \citep{Bayou} trained on a different form of program sketches that abstracted names and operations by their type. DreamCoder \citep{DREAMCODER} iteratively built sketches using progressively more complicated primitives though a wake-sleep algorithm. Our work is closely related in spirit but fundamentally differs in two ways: (1) our sketches are comprised of a general latent vocabulary that is learned in a simple, self-supervised fashion, and (2) our method avoids enumerative search, which is prohibitively expensive for large program spaces.
Another related avenue of research is using idiom mining to learn high-level concepts of a program \citep{idiom2019_1,idiom2019_2}. 
However, the idioms considered are always based on syntactic structure, i.e. subgraphs of the AST of the program, whereas tokens of our latent codes need not be so localized; also, idioms are extracted by a preprocessing step, whereas our training method learns the semantics of the latent tokens end-to-end. Finally, there is a line of work that deals with learning to process partial programs in addition to the specification. In \emph{execution-guided program synthesis}, the model guides iterative extensions of the partial programs until a matching one is found \citep{GARBAGECOLLECTOR,ChenLS19,REPL}. \citet{NeuralFixer} proposed training a differentiable fixer to edit incorrect programs. We treat these works as complementary, and can be combined with ours to refine predictions.




\textbf{Discrete Autoencoders.} Variational autoencoders (VAE) were first introduced using continuous latent representations \citep{VAE,rezende2014stochastic}. Several approaches were proposed to use discrete latent codes instead, such as continuous relaxations of categorical distributions i.e. the Gumbel-Softmax reparametrization trick \citep{GUMBELSOFTMAX,CONCRETE}. VQ-VAEs (see \autoref{sec:background_vq} for more details) achieved impressive results almost matching continuous VAEs ~\citep{VQVAE,Roy2018}.
In natural language processing, discrete bottlenecks have also been used for sentence compression \citep{LanguageLatentVariable}
and text generation \citep{Puduppully2019}, but these works do not use an autoencoder to learn the semantics of the latent codes, like our work does. 
Within the domain of synthesis of chemical molecules, \citep{Gomez-Bombarelli2018-gr}
have applied Bayesian optimization within a continuous latent space to guide this
structured prediction problem.
Learning to search has also been considered 
in the structured prediction literature \citep{Searn,BetterThanTeacher,Dagger},
but to our knowledge, these works do not consider the problem of learning
a discrete representation for search.
Notably, VQ-VAE methods have been successfully used to encode natural language into discrete codes for faster decoding in machine translation \citep{Kaiser2018}. The key novelty behind our work is in proposing two-level search over a learned latent discrete space; using a VQ-VAE as \citet{Kaiser2018} did enabled us to do so.

\section{Experiments}

We now present the results of evaluating our Latent Programmer model in two test domains: synthesis of string transformation programs from examples and code generation from natural language descriptions. We compare our LP model against several strong baselines. 

\textbf{RobustFill [LSTM]} is a seq-to-seq LSTM with attention on the input specification, and trained to autoregressively predict the true program. The architecture is comparable to the RobustFill model designed originally for the string transformation tasks in our first domain \citep{ROBUSTFILL}, but easily generalizes to all program synthesis domains. We detail the architecture in \autoref{sec:extended_robustfill}.

\textbf{RobustFill [Transformer]} alternatively uses a Transformer architecture, equivalent in architecture to the latent planner in our LP model, also trained to autoregressively predict the program. Transformers were found to perform much better than LSTMs in language tasks because they process the entire input as a whole, and have no risk of forgetting past dependencies \citep{Transformer}. This baseline can be also be considered an ablation of LP without latent codes.

To test the hypothesis that the discrete latent code is helping to organize search,
rather than simply increasing the capacity of the model, we compare to two ablations that use continuous autoencoders
rather than discrete ones. Because for these methods the latent space is continuous, 
combinatorial search algorithms such as beam search cannot search over the latent space.

\textbf{Latent RobustFill [AE]} replaces the VQ-VAE component of our LP model with a generic autoencoder. This makes the latent code a sequence of continuous embeddings. The latent prediction loss in \eqref{eq:lp_loss} is simply replaced by a squared error between the output of the autoencoder and the latent predictor. Performing beam search over the continuous latent space is intractable, so during inference we generate only one latent code per task; this is equivalent to two-level beam search described earlier with $L=1$. In addition, because we cannot define an end-of-sequence token in the latent space, this baseline must be given knowledge of the true program length even during inference, and always generates a latent code of length $\lceil T / 2^\ell \rceil$. 

\textbf{Latent RobustFill [VAE]} substitutes the VQ-VAE component with a VAE \citep{VAE}. This again produces a continuous latent space, but regularized to be distributed approximately as a standard Gaussian. Performing beam search is still intractable, but we can sample $L$ latent codes from the output of the VAE, and perform beam search on the programs afterwards. Again, we assume that the true program length is known during inference. 

\vspace{-0.05in}
\subsection{String Transformation}
\vspace{-0.05in}

The first test domain is a string transformation DSL frequently studied in the program synthesis literature \citep{NeuralProgramSynthesis,ROBUSTFILL,NeuralFixer}. Tasks in this domain involve finding a program which maps
a set of input strings to a corresponding set of outputs. Programs in the DSL are a concatenation of expressions that perform regex-based string transformations (see \autoref{sec:extended_robustfill}).

\begin{table}
\centering
\begin{tabular}{@{} p{4cm} p{0.75cm} p{0.75cm} p{0.75cm}@{}}
 \toprule
 \multirow{2}{*}{Method} & \multicolumn{3}{c}{Accuracy} \\
 \cmidrule{2-4}
 & B = 1 & 10 & 100 \\
 \midrule
 RobustFill [LSTM]  & 45$\%$ & 49$\%$ & 61$\%$ \\
 RobustFill [Transformer]  & 47$\%$ & 51$\%$ & 61$\%$ \\
 Latent RobustFill [AE] & 47$\%$ & 50$\%$ & 60$\%$ \\
 Latent RobustFill [VAE] & 46$\%$ & 51$\%$ & 62$\%$ \\
 Latent Programmer & \textbf{51}$\%$ & \textbf{57}$\%$ & \textbf{68}$\%$ \\
 \bottomrule
\end{tabular}
\vspace{-1mm}
\caption{Accuracy on string transformation domain.}
\label{table:robustfill_acc}
\vspace{-3mm}
\end{table}
We perform experiments on a synthetic dataset generated by sampling programs from the DSL, then the corresponding I/O examples using an heuristic 
similar to the one used in NSPS \citep{NeuralProgramSynthesis} and RobustFill \citep{ROBUSTFILL} to ensure nonempty output for each input.
We consider programs comprising of a concatenation of up to 10 expressions
and limit the lengths of strings in the I/O to be at most 100 characters. 
All models have an embedding size of 128 and hidden size of 512, and the attention layers consist of 3 stacked layers with 4 heads each. For the LP model, we used a latent compression factor $\ell=2$ and vocabulary size $K=40$.
The models are trained on roughly 25M tasks, and evaluated on 1K held-out  ones. 

In \autoref{table:robustfill_acc}, we report the accuracy --- the number of times a program was found conforming to the I/O examples --- of our method against the baselines. Across all beam sizes, our LP model performed 5-7 percentage points better (over 10$\%$ of baseline accuracy) than the next best model. From our ablative study, we see that having two-level using discrete latent codes was important, as the baselines over continuous latent spaces performed comparably to baseline RobustFill.

\begin{table}
\centering
\begin{tabular}{@{} p{5.5cm} p{1.5cm}@{}}
 \toprule
 Method & Accuracy \\
 \midrule
 DeepCoder  \citep{DEEPCODER} & 40$\%$\\
 SketchAdapt  \citep{SKETCHADAPT} & 62$\%$ \\
 Latent Programmer & \textbf{67}$\%$ \\
 \bottomrule
\end{tabular}
\vspace{-1mm}
\caption{Accuracy on string transformation domain of \citet{SKETCHADAPT} using $B=100$. \textsc{SketchAdapt} and \textsc{DeepCoder} results are from \citet{SKETCHADAPT} using $3,000$ and $300,000$ synthesized programs, respectively (similar wall clock time).} 
\label{table:neural_sketch_acc}
\vspace{-2mm}
\end{table}
\textbf{SketchAdapt } As alluded to earlier, two-level search was also proposed by \citet{SKETCHADAPT} as \textsc{SketchAdapt}, which learned programs with a \texttt{HOLE} token, then filled in the holes using enumerative search. To compare our proposed method with \textsc{SketchAdapt}, we evaluate our LP model on samples generated according to \citet{SKETCHADAPT}, which slightly modifies the DSL to improve the performance of synthesizers. We report results in \autoref{table:neural_sketch_acc}. Since enumeration can be done more quickly than beam search, we let \textsc{SketchAdapt} synthesize $3,000$ programs using $B=100$ top-level beams, whereas our LP model can only generate $B$ programs. We also reported results for \textsc{DeepCoder} \citep{DEEPCODER}, which synthesizes $300,000$ programs without the high-level beam search. We chose the number of synthesized programs so that all methods have similar wall clock time. Our LP model is able to outperform both methods in this modified DSL.

\vspace{-0.05in}
\subsection{Analysis}
\vspace{-0.05in}
\label{sec:analysis}

We conduct extensive analysis to better understand our LP model, the ability to generate long programs, and diversity in the beams. All results are reported with beam size $B = 10$. 

\begin{figure}[tb!]
\centering
\includegraphics[width=\linewidth]{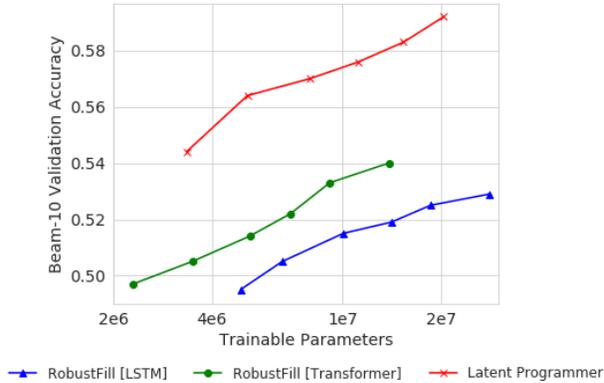}
\caption{Influence of hidden size on beam-10 accuracy.}
\label{fig:robustfill_size}
\end{figure}
\vspace{-0.05in}\paragraph{Model Size} Our LP model uses an additional latent code for decoding, which introduces additional parameters into the model than the baseline RobustFill model. To make a fair comparison, we vary the embedding and hidden dimension of all of our evaluated methods, and compare the effect of the number of trainable parameters on the accuracy. \autoref{fig:robustfill_size} shows that all methods respond well to an increase in model size. Nevertheless, we see that even when normalized for size, our LP model significantly outperforms baselines. 

\begin{table}[tb!]
    \centering
    \begin{tabular}{@{}c c c@{}}
     \toprule
     Length & RobustFill Acc. & LP Acc. \\
     \midrule
     1 & \textbf{94.5}$\%$ &  94.0$\%$ \\
     2 & 83.9$\%$ &  \textbf{84.6}$\%$ \\
     3 & \textbf{72.8}$\%$ &  72.2$\%$ \\
     4 & 63.1$\%$ &  \textbf{66.1}$\%$ \\
     5 & 47.1$\%$ &  \textbf{49.8}$\%$ \\
     6 & 40.6$\%$ &  \textbf{43.0}$\%$ \\
     7 & 30.2$\%$ &  \textbf{34.6}$\%$ \\
     8 & 22.7$\%$ &  \textbf{28.4}$\%$ \\
     9 & 18.6$\%$ &  \textbf{27.0}$\%$ \\
     10 & 14.4$\%$ & \textbf{25.6}$\%$ \\
    \bottomrule
    \end{tabular}
    \caption{Beam-10 accuracy of baseline transformer and LP by ground truth program length.}
    \label{tab:robustfill_length}
\vspace{-0.1in}
\end{table}
\vspace{-0.05in}\paragraph{Program Length} Prior work has shown that program length is a reasonable proxy measure of problem difficulty. We hypothesize that using latent codes is most beneficial when generating long programs. \autoref{tab:robustfill_length} shows how ground-truth program length affects the accuracy of our LP model compared to RobustFill, which lacks latent codes. As expected, accuracy decreases with problem complexity. Perhaps surprisingly, though, we see a large improvement in our LP model's ability to handle more complex problems. This supports our hypothesis two-level search can organize and improve search over more complex tasks, because we see a greater improvement in accuracy precisely for the examples in which traditional search is most difficult.
\begin{figure*}[t]
\centering
\begin{minipage}{\textwidth}
    \centering
    \begin{tabular}{@{}l l l@{}}
     \toprule
     Inputs & Outputs & Program \\
     \midrule
     ``Jacob,Ethan,James 11" &  ``11:J.E.J." &  \small\texttt{GetToken\_NUMBER\_1 \ \ | Const(:) |}\\
     ``Elijah,Daniel,Aiden 3162" & ``3162:E.D.A" & \small\texttt{GetToken\_ALL\_CAPS\_1 | Const(.) |} \\
     ``Rick,Oliver,Mia 26" & ``26:R.O.M." & \small\texttt{GetToken\_ALL\_CAPS\_2 | Const(.) |}\\
     ``Mark,Ben,Sam 510" & ``510:M.B.S." & \small\texttt{GetToken\_ALL\_CAPS\_3 | Const(.)} \\
    \bottomrule
    \end{tabular}
\end{minipage}
\begin{minipage}{\textwidth}
    \vspace{0.1in}
    \centering
    \begin{tabular}{p{1.5cm} | p{12.5cm}}
    RobustFill &  \small\texttt{GetAll\_NUMBER | Const(:)| GetToken\_ALL\_CAPS\_2 | Const(.)} \\\rule{0pt}{3ex}
    LP & \small\texttt{GetAll\_NUMBER | Const(:) | GetToken\_ALL\_CAPS\_1 | Const(.) | GetToken\_ALL\_CAPS\_2 | Const(.) | GetToken\_ALL\_CAPS\_-1 | Const(.)} \\\rule{0pt}{3ex}
    LP Latent & \small\texttt{TOK\_14 | TOK\_36 | TOK\_36 | TOK\_36} \\
    \end{tabular}
\end{minipage}
\caption{Illustrative string transformation problem where the ground-truth program was long but had repetitive structure. The baseline Transformer was unable to generate the program but our LP model, which first predicts a coarse latent code, was able to.}
\label{fig:robustfill_example_2}
\end{figure*}

\vspace{-0.05in}\paragraph{Latent Beam Size} In two-level beam search of beam size $B$, first $L$ latent beams are decoded, then $\lfloor B / L \rfloor$ programs per latent code. The latent beam size $L$ controls how much search is performed over latent space. We theorize that higher $L$ will produce more diverse beams; however, too high $L$ can be harmful in missing programs with high joint log-probability. We show the effect of latent beam size on both the beam-10 accuracy and a proxy measure for diversity.
Diversity is important to measure because increased diversity suggests that two-level search is better exploring the space of possible programs.
Following prior work, we measure diversity by counting the number of distinct $n$-grams in the beams, normalized by the total number of tokens to bias against long programs \citep{DiverseBeamSearch}.  We report the results varying $L$ for $B=10$ in \autoref{tab:robustfill_diversity}. As expected, increasing the latent beam size $L$ improves diversity of output programs, but excessively large $L$ harms the final accuracy. An important observation is that the $L=1$ case effectively corresponds to single-level search, and performs similarly to baseline RobustFill. This is further evidence that explicitly having two-level search is critical to the LP model's improved performance.

\begin{table}
\centering
\begin{tabular}{@{} c c  c c c c @{}}
 \toprule
 \multirow{2}{*}{Beam Size} & \multirow{2}{*}{Accuracy} & \multicolumn{4}{c}{Distinct n-Grams} \\
 \cmidrule{3-6}
 & & n = 1 & 2 & 3 & 4 \\
 \midrule
 L = 1  & 52$\%$ & 0.13 & 0.23 & 0.26 & 0.28\\
 2 & 55$\%$ & 0.13 & 0.24 & 0.26 & 0.28 \\
 3 & \textbf{57}$\%$ & 0.14 & 0.25 & 0.28 & 0.31 \\
 5 & 57$\%$& 0.14 & 0.26 & 0.29 & 0.32 \\
 10 & 56$\%$ & \textbf{0.14} & \textbf{0.26} & \textbf{0.30} & \textbf{0.33} \\
 \bottomrule
\end{tabular}
\caption{Effect of latent beam size on beam-10 accuracy and number of distinct $n$-grams (normalized by total number of tokens).}
\label{tab:robustfill_diversity}
\vspace{-0.1in}
\end{table}

\setlength{\columnsep}{6pt}%
\begin{wrapfigure}{l}{0.38\linewidth}
\begin{minipage}{\linewidth}
\vspace{-0.15in}
\centering
\begin{tabular}{@{} l c @{}}
 \toprule
 $2^\ell$ & Accuracy \\
 \midrule
 2\phantom{10} & 52$\%$\\
 4 & \textbf{55}$\%$\\
 8 & 49$\%$\\
\bottomrule
\end{tabular}
\end{minipage}
\begin{minipage}{\linewidth}
\centering
\begin{tabular}{@{} l c @{}}
 \toprule
 $K$ & Accuracy \\
 \midrule
 10 & 48$\%$\\
 40 & \textbf{55}$\%$\\
 100 & 51$\%$\\
\bottomrule
\end{tabular}
\end{minipage}
\caption{Effect of  $\ell, K$.}
\label{fig:robustfill_latent_hyperparameters}
\vspace{-0.1in}
\end{wrapfigure}
\vspace{-0.05in}\paragraph{Latent Code Dimension} We also measured the effect of the expressiveness of our latent code, specifically by varying the latent length compression factor $\ell$,
and size of latent vocabulary $K$, on overall performance.
If $c$ is too small, the latent space becomes too large to search; on the other hard, too large $c$ can mean individual latent tokens cannot encode enough information to reconstruct the program. Similarly, we expect that too small of a vocabulary $K$ can limit the expressiveness of the latent space, but too large $K$ can make the latent space too complex, and predicting the correct latent code difficult. \autoref{fig:robustfill_latent_hyperparameters} confirms this. 

\begin{figure*}[t]
    \centering
    \begin{tabular}{@{}l c c c c c c c@{}}
     \toprule
      & \texttt{TOK\_3} & \texttt{TOK\_4} & \texttt{TOK\_5} & \texttt{TOK\_6} & \texttt{TOK\_7} & \texttt{TOK\_8} & \texttt{TOK\_9} \\
     \midrule
     Get First Number & 12$\%$ & 5$\%$ & 0$\%$ & 9$\%$ & 70$\%$ & 6$\%$ & 0$\%$ \\
     Get Last Number & 22$\%$ & 49$\%$ & 0$\%$ & 11$\%$ & 8$\%$ & 8$\%$ & 0$\%$ \\
     Get First Word & 10$\%$ & 20$\%$ & 0$\%$ & 56$\%$ & 7$\%$ & 9$\%$ & 0$\%$ \\
     Get Last Word & 75$\%$ & 4$\%$ & 0$\%$ & 6$\%$ & 9$\%$ & 6$\%$ & 0$\%$ \\
     Get First Alphanum & 11$\%$ & 3$\%$ & 0$\%$ & 35$\%$ & 42$\%$ & 9$\%$ & 0$\%$ \\
     Get Last Alphanum & 45$\%$ & 29$\%$ & 0$\%$ & 22$\%$ & 0$\%$ & 4$\%$ & 0$\%$ \\
    \bottomrule
    \end{tabular}
    \caption{Percentage of time each high-level operation was associated with a particular latent token on toy DSL. Note that tokens $0, 1, 2$ are reserved for padding, and start and end of sequences, respectively.}
    \label{fig:co_occurence_table}
\end{figure*}
\vspace{-0.05in}\paragraph{Latent Interpretability} A key hypothesis of our work is that searching over latent codes organizes search over programs; it is crucial that the latent codes be informative of the synthesized program. 
In \autoref{fig:robustfill_example_2}, we also show an illustrative example in the domain where our LP model found a valid program whereas the RobustFill model did not (more examples are in \autoref{sec:supp_examples}). In the example, the ground-truth program was long but had a repetitive underlying structure. Our LP model correctly detected this structure, as evidenced by the predicted latent code. However, due to the complexity of our DSL and size of latent space, it is difficult to find explicit meaning behind individual tokens.

Thus, to better investigate interpretability, we created a toy DSL using only the \T{GetSpan} expression from the RobustFill DSL. This expression allows us to grab arbitrary ranges defined by a regex and its index of appearance, so sufficiently complex programs can still be generated (see \autoref{sec:interpretability} for full DSL). 
We trained a LP model with $\ell = 2$ and $K = 10$ on the toy DSL, and recorded examples of predicted programs and their corresponding latent codes in \autoref{sec:interpretability}. From these examples, we can see a pattern of LP associating particular latent tokens with high-level operations. For example, \texttt{TOK\_7} and \texttt{TOK\_4} were extracting the first and last number in the string, and \texttt{TOK\_6}, \texttt{TOK\_3} the first and last word. As further evidence, in \autoref{fig:co_occurence_table}, we chose six high-level operations and recorded the percentage of times each was mapped to a specific latent token. Specific high-level operations were clearly biased towards particular latent tokens, further suggesting that the latent codes were specifying high-level components of the program. In addition, since there are multiple syntactic ways of expressing the same operation, latent tokens were more likely capturing high-level semantics over syntax.

\vspace{-0.05in}
\subsection{Python Code Generation}
\vspace{-0.05in}

\begin{figure*}[t]
\centering
\begin{tabular}{@{}l l l@{}}
 \toprule
 Docstring & Program\\[0.1cm]
 \midrule
 return a list of the words & \small\texttt{def split(s, sep=None, maxsplit=-1):} \\
 in the string s & \quad\small\texttt{return s.split(sep, maxsplit)} \\
\bottomrule
\end{tabular}
\caption{Example problem from the Python code generation dataset.}
\label{fig:python_example}
\end{figure*}

Our next test domain is a Python code generation (CG) task, which involves generating code for a function that implements a natural-language specification. The dataset used consists of 111K python examples, which consist of a docstring and corresponding code snippet, collected from Github \citep{PythonCode}. An example docstring and program from the dataset is shown in \autoref{fig:python_example}. 

We used a language-independent tokenizer jointly on data \citep{Sentencepiece}, and processed the dataset into a vocabulary of 35K sub-word tokens. Furthermore, following \citet{DualPythonCode}, we set the maximum length of the programs to be 150 tokens resulting in 85K examples. 
Across all models, we set the embedding size to be 256 and hidden size to be 512, and the attention layers consist of 6 stacked layers with 16 heads each, similar to in neural machine translation \citep{Transformer}. For the LP model, we used a latent compression factor $c = 2$ and vocabulary size $K = 400$ after a hyperparameter search.
The models are evaluated on 1K held-out examples. 
We initially found that it was difficult for the program encoder to find a latent structure in the ground-truth programs due to the 
wide variety of variable names. To remedy this,
we replace the $i$-th function argument and variable appearing the program with the token \texttt{ARG\_i} and \texttt{VAR\_i}, respectively. This was only used in training the program encoder.

\begin{table}
\centering
\begin{tabular}{@{}p{4cm} p{0.75cm} p{0.75cm} p{0.75cm}@{}}
 \toprule
 \multirow{2}{*}{Method} & \multicolumn{3}{c}{BLEU} \\
 \cmidrule{2-4}
 & B = 1 & 10 & 100 \\
 \midrule
 Base \citep{DualPythonCode} & 10.4 & - & - \\
 Dual \citep{DualPythonCode} & 12.1 & - & - \\
 \midrule
 RobustFill [LSTM]   & 11.4 & 14.8 & 16.0 \\
 RobustFill [Transformer]  & 12.1 & 15.5 & 17.2 \\
 Latent Programmer & \textbf{14.0} & \textbf{18.6} & \textbf{21.3} \\
 \bottomrule
\end{tabular}
\caption{BLEU score on code generation task.}
\label{table:python_accuracy}
\vspace{-0.1in}
\end{table}

In this domain, we are not given I/O examples as specification. In addition, the programs in the dataset are often not executable due missing dependencies, or having complex objects as arguments. Hence, we cannot measure accuracy by evaluating programs on test cases as before. Instead, we evaluate performance by computing the best BLEU score among the output beams \citep{BLEU}. This is a natural metric, as we can imagine that in practice, a user would examine the  candidate programs to select one that best matches their intent. We computed BLEU as the geometric mean of $n$-gram matching precision scores up to $n = 4$.  \autoref{table:python_accuracy} shows that our LP model outperforms the baselines. 
From the results, it can be seen that this is a difficult task, which may be due to the ambiguity in specifying code from a short docstring description.
As evidence, we additionally include results from a recent work that proposed seq-to-seq CG models on the same data that performed similar to our baselines \citep{DualPythonCode}. These results show that improvements due to the LP model exist even in difficult CG domains. For example docstrings and generated code, refer to \autoref{sec:supp_examples}.

\begin{figure*}[t]
\centering
\begin{tabular}{|l|c c c c c|}
\hline
0&    \texttt{\_files} & \texttt{dirname} & \texttt{glob} & \texttt{isdir} & \texttt{makedir} \\
\hline
1&    \texttt{server} & \texttt{\_port} & \texttt{\_socket} & \texttt{\_password} & \texttt{host} \\
\hline
2&  \texttt{pip} & \texttt{package} & \texttt{wheel} & \texttt{install} & \texttt{sudo} \\
\hline
3&  \texttt{dt} & \texttt{interval} & \texttt{seconds} & \texttt{time} & \texttt{timestamp} \\
\hline
4&  \texttt{timeout} & \texttt{\_timeout} & \texttt{handle} & \texttt{future} & \texttt{notifier} \\
\hline
\end{tabular}
\caption{Example latent tokens and top-5 program tokens ranked by TF-IDF score.}
\label{fig:python_interpretability}
\vspace{-0.1in}
\end{figure*}
Finally, we investigated interpretability in this domain. For each latent token, we collected the set of programs associated with that token, and for each of those sets, we ranked the program tokens by TF-IDF \citep{salton1986introduction}. 
In \autoref{fig:python_interpretability}, we list several latent tokens where the top-5 program tokens have a common semantic interpretation.
For example, the first one seems to exhibit a latent state learning high-level concepts about file manipulation. However, due to the scale and noisiness of the dataset, it was difficult to see strong semantic clustering among all latent tokens.

\section{Conclusion}
In this work we proposed the Latent Programmer (LP), a novel neural program synthesis technique that leverages a structured latent sequences to guide search. The LP model consists of a latent predictor, which maps the input specification to a sequence of discrete latent variables, and a latent program decoder that generates a program token-by-token while attending to the latent sequence.   
The latent predictor was trained via a self-supervised method in which a discrete autoencoder of programs was learned using a discrete bottleneck, specifically a VQ-VAE ~\citep{VQVAE}, and the latent predictor tries to predict the autoencoded sequence as if it were the ground-truth. During inference, the LP model first searches in latent space for discrete codes, then conditions on those codes to search over programs. 
Empirically, we showed that the Latent Programmer outperforms state-of-the-art baselines as Robustfill ~\citep{ROBUSTFILL}, which ignore latent structure. Exciting future avenues of investigation include achieving better performance by grounding the latent vocabulary and generalizing our method to other tasks in structured prediction.

\bibliographystyle{icml2021}
\bibliography{latent_programmer}

\begin{thebibliography}{46}
\providecommand{\natexlab}[1]{#1}
\providecommand{\url}[1]{\texttt{#1}}
\expandafter\ifx\csname urlstyle\endcsname\relax
  \providecommand{\doi}[1]{doi: #1}\else
  \providecommand{\doi}{doi: \begingroup \urlstyle{rm}\Url}\fi

\bibitem[Alur et~al.(2013)Alur, Bod{\'{\i}}k, Juniwal, Martin, Raghothaman,
  Seshia, Singh, Solar{-}Lezama, Torlak, and Udupa]{sygus}
Alur, R., Bod{\'{\i}}k, R., Juniwal, G., Martin, M. M.~K., Raghothaman, M.,
  Seshia, S.~A., Singh, R., Solar{-}Lezama, A., Torlak, E., and Udupa, A.
\newblock Syntax-guided synthesis.
\newblock In \emph{Formal Methods in Computer-Aided Design, {FMCAD} 2013,
  Portland, OR, USA, October 20-23, 2013}, pp.\  1--8. {IEEE}, 2013.

\bibitem[Bahdanau et~al.(2016)Bahdanau, Cho, and Bengio]{Attention}
Bahdanau, D., Cho, K., and Bengio, Y.
\newblock Neural machine translation by jointly learning to align and
  translate.
\newblock In \emph{International Conference on Learning Representations
  (ICLR)}, 2016.

\bibitem[Bahuleyan et~al.(2017)Bahuleyan, Mou, Vechtomova, and
  Poupart]{Bypassing}
Bahuleyan, H., Mou, L., Vechtomova, O., and Poupart, P.
\newblock Variational attention for sequence-to-sequence models.
\newblock \emph{CoRR}, abs/1712.08207, 2017.

\bibitem[Balog et~al.(2017)Balog, Gaunt, Brockschmidt, Nowozin, and
  Tarlow]{DEEPCODER}
Balog, M., Gaunt, A.~L., Brockschmidt, M., Nowozin, S., and Tarlow, D.
\newblock Deepcoder: Learning to write programs.
\newblock In \emph{International Conference on Learning Representations
  (ICLR)}, 2017.

\bibitem[Balog et~al.(2020)Balog, Singh, Maniatis, and Sutton]{NeuralFixer}
Balog, M., Singh, R., Maniatis, P., and Sutton, C.
\newblock Neural program synthesis with a differentiable fixer.
\newblock \emph{CoRR}, abs/2006.10924, 2020.
\newblock URL \url{https://arxiv.org/abs/2006.10924}.

\bibitem[Barke et~al.(2020)Barke, Peleg, and Polikarpova]{Barke2020-rq}
Barke, S., Peleg, H., and Polikarpova, N.
\newblock {Just-in-Time} learning for {Bottom-Up} enumerative synthesis.
\newblock In \emph{Object-oriented Programming, Systems, Languages, and
  Applications ({OOPSLA})}, 2020.

\bibitem[Chang et~al.(2015)Chang, Krishnamurthy, Agarwal, {Daume III}, and
  Langford]{BetterThanTeacher}
Chang, K.-W., Krishnamurthy, A., Agarwal, A., {Daume III}, and Langford, J.
\newblock Learning to search better than your teacher.
\newblock In \emph{International Conference on Machine Learning ({ICML})},
  2015.

\bibitem[Chen et~al.(2019)Chen, Liu, and Song]{ChenLS19}
Chen, X., Liu, C., and Song, D.
\newblock Execution-guided neural program synthesis.
\newblock In \emph{International Conference on Learning Representations
  (ICLR)}, 2019.

\bibitem[Daum{\'e} et~al.(2009)Daum{\'e}, Langford, and Marcu]{Searn}
Daum{\'e}, III, H., Langford, J., and Marcu, D.
\newblock Search-based structured prediction.
\newblock \emph{Machine Learning Journal}, 2009.

\bibitem[Devlin et~al.(2017)Devlin, Uesato, Bhupatiraju, Singh, Mohamed, and
  Kohli]{ROBUSTFILL}
Devlin, J., Uesato, J., Bhupatiraju, S., Singh, R., Mohamed, A., and Kohli, P.
\newblock Robustfill: Neural program learning under noisy {I/O}.
\newblock \emph{CoRR}, abs/1703.07469, 2017.
\newblock URL \url{http://arxiv.org/abs/1703.07469}.

\bibitem[Ellis et~al.(2019)Ellis, Nye, Pu, Sosa, Tenenbaum, and
  Solar{-}Lezama]{REPL}
Ellis, K., Nye, M.~I., Pu, Y., Sosa, F., Tenenbaum, J., and Solar{-}Lezama, A.
\newblock Write, execute, assess: Program synthesis with a {REPL}.
\newblock In \emph{Neural Information Processing Systems (NeurIPS)}, 2019.

\bibitem[Ellis et~al.(2020)Ellis, Wong, Nye, Sable-Meyer, Cary, Morales,
  Hewitt, Solar-Lezama, and Tenenbaum]{DREAMCODER}
Ellis, K., Wong, C., Nye, M., Sable-Meyer, M., Cary, L., Morales, L., Hewitt,
  L., Solar-Lezama, A., and Tenenbaum, J.~B.
\newblock Dreamcoder: Growing generalizable, interpretable knowledge with
  wake-sleep bayesian program learning.
\newblock \emph{CoRR}, abs/2006.08381, 2020.
\newblock URL \url{https://arxiv.org/abs/2006.08381}.

\bibitem[G{\'o}mez-Bombarelli et~al.(2018)G{\'o}mez-Bombarelli, Wei, Duvenaud,
  Hern{\'a}ndez-Lobato, S{\'a}nchez-Lengeling, Sheberla, Aguilera-Iparraguirre,
  Hirzel, Adams, and Aspuru-Guzik]{Gomez-Bombarelli2018-gr}
G{\'o}mez-Bombarelli, R., Wei, J.~N., Duvenaud, D., Hern{\'a}ndez-Lobato,
  J.~M., S{\'a}nchez-Lengeling, B., Sheberla, D., Aguilera-Iparraguirre, J.,
  Hirzel, T.~D., Adams, R.~P., and Aspuru-Guzik, A.
\newblock Automatic chemical design using a {Data-Driven} continuous
  representation of molecules.
\newblock \emph{ACS Cent Sci}, 4\penalty0 (2):\penalty0 268--276, February
  2018.

\bibitem[Gulwani(2011)]{FLASHFILL}
Gulwani, S.
\newblock Automating string processing in spreadsheets using input-output
  examples.
\newblock In \emph{PoPL'11, January 26-28, 2011, Austin, Texas, USA}, 2011.

\bibitem[Gulwani et~al.(2017)Gulwani, Polozov, and Singh]{now17}
Gulwani, S., Polozov, O., and Singh, R.
\newblock Program synthesis.
\newblock \emph{Foundations and Trends in Programming Languages}, 4\penalty0
  (1-2):\penalty0 1--119, 2017.
\newblock \doi{10.1561/2500000010}.
\newblock URL \url{https://doi.org/10.1561/2500000010}.

\bibitem[Iyer et~al.(2019)Iyer, Cheung, and Zettlemoyer]{idiom2019_2}
Iyer, S., Cheung, A., and Zettlemoyer, L.
\newblock Learning programmatic idioms for scalable semantic parsing.
\newblock In \emph{EMNLP-IJCNLP}, 2019.

\bibitem[Jang et~al.(2017)Jang, Gu, and Poole]{GUMBELSOFTMAX}
Jang, E., Gu, S., and Poole, B.
\newblock Categorical reparameterization with gumbel-softmax.
\newblock In \emph{International Conference on Learning Representations
  (ICLR)}, 2017.

\bibitem[Kaiser et~al.(2018)Kaiser, Roy, Vaswani, Parmar, Bengio, Uszkoreit,
  and Shazeer]{Kaiser2018}
Kaiser, {\L}., Roy, A., Vaswani, A., Parmar, N., Bengio, S., Uszkoreit, J., and
  Shazeer, N.
\newblock Fast decoding in sequence models using discrete latent variables.
\newblock In \emph{International Conference on Machine Learning (ICML)}, 2018.

\bibitem[Kingma \& Welling(2014)Kingma and Welling]{VAE}
Kingma, D.~P. and Welling, M.
\newblock Auto-encoding variational bayes.
\newblock In \emph{International Conference on Learning Representations
  (ICLR)}, 2014.

\bibitem[Koza(1994)]{koza1994genetic}
Koza, J.~R.
\newblock Genetic programming as a means for programming computers by natural
  selection.
\newblock \emph{Statistics and computing}, 4\penalty0 (2):\penalty0 87--112,
  1994.

\bibitem[Kudo \& Richardson(2018)Kudo and Richardson]{Sentencepiece}
Kudo, T. and Richardson, J.
\newblock {S}entence{P}iece: A simple and language independent subword
  tokenizer and detokenizer for neural text processing.
\newblock In \emph{Proceedings of the 2018 Conference on Empirical Methods in
  Natural Language Processing: System Demonstrations}, pp.\  66--71, November
  2018.

\bibitem[Lee et~al.(2018)Lee, Heo, Alur, and Naik]{EUPHONY}
Lee, W., Heo, K., Alur, R., and Naik, M.
\newblock Accelerating search-based program synthesis using learned
  probabilistic models.
\newblock In \emph{Conference on Programming Language Design and Implementation
  (PLDI)}, pp.\  436--449, June 2018.

\bibitem[Maddison et~al.(2017)Maddison, Mnih, and Teh]{CONCRETE}
Maddison, C.~J., Mnih, A., and Teh, Y.~W.
\newblock The concrete distribution: A continuous relaxation of discrete random
  variables.
\newblock In \emph{International Conference on Learning Representations
  (ICLR)}, 2017.

\bibitem[Manna \& Waldinger(1971)Manna and Waldinger]{mannaw71}
Manna, Z. and Waldinger, R.~J.
\newblock Toward automatic program synthesis.
\newblock \emph{Commun. {ACM}}, 14\penalty0 (3):\penalty0 151--165, 1971.

\bibitem[Miao \& Blunsom(2016)Miao and Blunsom]{LanguageLatentVariable}
Miao, Y. and Blunsom, P.
\newblock Language as a latent variable: Discrete generative models for
  sentence compression.
\newblock \emph{CoRR}, abs/1609.07317, 2016.
\newblock URL \url{http://arxiv.org/abs/1609.07317}.

\bibitem[Murali et~al.(2018)Murali, Qi, Chaudhuri, and Jermaine]{Bayou}
Murali, V., Qi, L., Chaudhuri, S., and Jermaine, C.
\newblock Neural sketch learning for conditional program generation.
\newblock In \emph{International Conference on Learning Representations
  ({ICLR})}, 2018.

\bibitem[Nye et~al.(2019)Nye, Hewitt, Tenenbaum, and
  Solar{-}Lezama]{SKETCHADAPT}
Nye, M.~I., Hewitt, L.~B., Tenenbaum, J.~B., and Solar{-}Lezama, A.
\newblock Learning to infer program sketches.
\newblock In \emph{International Conference on Machine Learning (ICML)}, 2019.

\bibitem[Odena et~al.(2020)Odena, Shi, Bieber, Singh, Sutton, and Dai]{BUSTLE}
Odena, A., Shi, K., Bieber, D., Singh, R., Sutton, C., and Dai, H.
\newblock {BUSTLE}: {Bottom-Up} program synthesis through learning-guided
  exploration.
\newblock In \emph{International Conference on Learning Representations
  ({ICLR})}, September 2020.

\bibitem[Papineni et~al.(2002)Papineni, Roukos, Ward, and Zhu]{BLEU}
Papineni, K., Roukos, S., Ward, T., and Zhu, W.-J.
\newblock Bleu: A method for automatic evaluation of machine translation.
\newblock In \emph{Proceedings of the 40th Annual Meeting on Association for
  Computational Linguistics}, pp.\  311–318. Association for Computational
  Linguistics, 2002.

\bibitem[Parisotto et~al.(2017)Parisotto, Mohamed, Singh, Li, Zhou, and
  Kohli]{NeuralProgramSynthesis}
Parisotto, E., Mohamed, A., Singh, R., Li, L., Zhou, D., and Kohli, P.
\newblock Neuro-symbolic program synthesis.
\newblock In \emph{International Conference on Learning Representations
  (ICLR)}, 2017.

\bibitem[Puduppully et~al.(2019)Puduppully, Dong, and Lapata]{Puduppully2019}
Puduppully, R., Dong, L., and Lapata, M.
\newblock Data-to-text generation with content selection and planning.
\newblock In \emph{The Thirty-Third {AAAI} Conference on Artificial
  Intelligence, {AAAI} 2019, The Thirty-First Innovative Applications of
  Artificial Intelligence Conference, {IAAI} 2019, The Ninth {AAAI} Symposium
  on Educational Advances in Artificial Intelligence, {EAAI} 2019, Honolulu,
  Hawaii, USA, January 27 - February 1, 2019}, pp.\  6908--6915. {AAAI} Press,
  2019.
\newblock \doi{10.1609/aaai.v33i01.33016908}.
\newblock URL \url{https://doi.org/10.1609/aaai.v33i01.33016908}.

\bibitem[Rezende et~al.(2014)Rezende, Mohamed, and
  Wierstra]{rezende2014stochastic}
Rezende, D.~J., Mohamed, S., and Wierstra, D.
\newblock Stochastic backpropagation and approximate inference in deep
  generative models.
\newblock \emph{CoRR}, abs/1401.4082, 2014.
\newblock URL \url{https://arxiv.org/abs/1401.4082}.

\bibitem[Ross et~al.(2011)Ross, Gordon, and Bagnell]{Dagger}
Ross, S., Gordon, G., and Bagnell, D.
\newblock A reduction of imitation learning and structured prediction to
  {No-Regret} online learning.
\newblock In Gordon, G., Dunson, D., and Dud{\'\i}k, M. (eds.),
  \emph{Conference on Artificial Intelligence and Statistics (AISTATS)},
  volume~15 of \emph{Proceedings of Machine Learning Research}, pp.\  627--635,
  Fort Lauderdale, FL, USA, 2011. PMLR.

\bibitem[Roy et~al.(2018)Roy, Vaswani, Neelakantan, and Parmar]{Roy2018}
Roy, A., Vaswani, A., Neelakantan, A., and Parmar, N.
\newblock Theory and experiments on vector quantized autoencoders.
\newblock \emph{arXiv}, May 2018.

\bibitem[Salton \& McGill(1986)Salton and McGill]{salton1986introduction}
Salton, G. and McGill, M.~J.
\newblock Introduction to modern information retrieval.
\newblock 1986.

\bibitem[Schkufza et~al.(2013)Schkufza, Sharma, and Aiken]{stoke}
Schkufza, E., Sharma, R., and Aiken, A.
\newblock Stochastic superoptimization.
\newblock In \emph{Proceedings of the Eighteenth International Conference on
  Architectural Support for Programming Languages and Operating Systems},
  ASPLOS ’13, pp.\  305–316, New York, NY, USA, 2013. Association for
  Computing Machinery.
\newblock ISBN 9781450318709.
\newblock \doi{10.1145/2451116.2451150}.
\newblock URL \url{https://doi.org/10.1145/2451116.2451150}.

\bibitem[Shin et~al.(2019)Shin, Allamanis, Brockschmidt, and
  Polozov]{idiom2019_1}
Shin, R., Allamanis, M., Brockschmidt, M., and Polozov, O.
\newblock Program synthesis and semantic parsing with learned code idioms.
\newblock In \emph{Neural Information Processing Systems (NeurIPS)}, 2019.

\bibitem[Solar{-}Lezama et~al.(2006)Solar{-}Lezama, Tancau, Bod{\'{\i}}k,
  Seshia, and Saraswat]{SKETCH}
Solar{-}Lezama, A., Tancau, L., Bod{\'{\i}}k, R., Seshia, S.~A., and Saraswat,
  V.~A.
\newblock Combinatorial sketching for finite programs.
\newblock In \emph{Conference on Architectural Support for Programming
  Languages and Operating Systems, {ASPLOS} 2006, San Jose, CA, USA, October
  21-25, 2006}, pp.\  404--415. {ACM}, 2006.

\bibitem[Summers(1977)]{summers1977methodology}
Summers, P.~D.
\newblock A methodology for lisp program construction from examples.
\newblock \emph{Journal of the ACM (JACM)}, 24\penalty0 (1):\penalty0 161--175,
  1977.

\bibitem[Udupa et~al.(2013)Udupa, Raghavan, Deshmukh, Mador-Haim, Martin, and
  Alur]{TRANSIT}
Udupa, A., Raghavan, A., Deshmukh, J.~V., Mador-Haim, S., Martin, M. M.~K., and
  Alur, R.
\newblock {TRANSIT}: Specifying protocols with concolic snippets.
\newblock In \emph{Conference on Programming Language Design and Implementation
  (PLDI)}, pp.\  287--296. Association for Computing Machinery, 2013.

\bibitem[van~den Oord et~al.(2017)van~den Oord, Vinyals, and
  Kavukcuoglu]{VQVAE}
van~den Oord, A., Vinyals, O., and Kavukcuoglu, K.
\newblock Neural discrete representation learning.
\newblock In \emph{Neural Information Processing Systems (NeurIPS)}, 2017.

\bibitem[Vaswani et~al.(2017)Vaswani, Shazeer, Parmar, Uszkoreit, Jones, Gomez,
  Kaiser, and Polosukhin]{Transformer}
Vaswani, A., Shazeer, N., Parmar, N., Uszkoreit, J., Jones, L., Gomez, A.~N.,
  Kaiser, L., and Polosukhin, I.
\newblock Attention is all you need.
\newblock In \emph{Neural Information Processing Systems (NeurIPS)}, 2017.

\bibitem[Vijayakumar et~al.(2018)Vijayakumar, Cogswell, Selvaraju, Sun, Lee,
  Crandall, and Batra]{DiverseBeamSearch}
Vijayakumar, A.~K., Cogswell, M., Selvaraju, R.~R., Sun, Q., Lee, S., Crandall,
  D.~J., and Batra, D.
\newblock Diverse beam search: Decoding diverse solutions from neural sequence
  models.
\newblock In \emph{AAAI}, 2018.

\bibitem[{Wan} et~al.(2018){Wan}, {Zhao}, {Yang}, {Xu}, {Ying}, {Wu}, and
  {Yu}]{PythonCode}
{Wan}, Y., {Zhao}, Z., {Yang}, M., {Xu}, G., {Ying}, H., {Wu}, J., and {Yu},
  P.~S.
\newblock Improving automatic source code summarization via deep reinforcement
  learning.
\newblock In \emph{2018 33rd IEEE/ACM International Conference on Automated
  Software Engineering (ASE)}, pp.\  397--407, 2018.

\bibitem[Wei et~al.(2019)Wei, Li, Xia, Fu, and Jin]{DualPythonCode}
Wei, B., Li, G., Xia, X., Fu, Z., and Jin, Z.
\newblock Code generation as a dual task of code summarization.
\newblock In \emph{Neural Information Processing Systems (NeurIPS)}, 2019.

\bibitem[Zohar \& Wolf(2018)Zohar and Wolf]{GARBAGECOLLECTOR}
Zohar, A. and Wolf, L.
\newblock Automatic program synthesis of long programs with a learned garbage
  collector.
\newblock In \emph{Neural Information Processing Systems (NeurIPS)}, 2018.

\end{thebibliography}

\clearpage
\onecolumn
\appendix

\section{Extended Description of DSL and RobustFill Model}
\label{sec:extended_robustfill}
\begin{figure}[ht]
\small
\begin{alignat*}{2}
\mbox{Program } Y\quad &:= &\quad& \T{Concat}(e_1, e_2, \hdots) \\
\mbox{Expression } e\quad &:= && f \logicalOR n \logicalOR n_1(n_2) \logicalOR n(f) \logicalOR \T{ConstStr}(c) \\ 
\mbox{Substring } f\quad &:= && \T{SubStr}(k_1, k_2) \logicalOR 
\T{GetSpan}(r_1, i_1, b_1, r_2, i_2, b_2)  \\
\mbox{Nesting } n\quad &:= && \T{GetToken}(t, i) \logicalOR  \T{ToCase}(s) \logicalOR \T{Replace}(\delta_1, \delta_2) \logicalOR \T{Trim}() \logicalOR \T{GetUpto}(r) 
\logicalOR \T{GetFrom}(r) \\
&&& \logicalOR \T{GetFirst}(t, i) \logicalOR \T{GetAll}(t) \\
\mbox{Regex } r\quad &:= && t_1 \logicalOR \hdots \logicalOR t_n \logicalOR \delta_1 \logicalOR \hdots \logicalOR \delta_m \\
\mbox{Type } t\quad &:= &&  \T{NUMBER} \logicalOR \T{WORD} \logicalOR \T{ALPHANUM} \logicalOR \T{ALL\_CAPS} \logicalOR \T{PROP\_CASE} \logicalOR \T{LOWER} \logicalOR \T{DIGIT} \logicalOR \T{CHAR} \\
\mbox{Case } s\quad &:= && \T{PROPER} \logicalOR \T{ALL\_CAPS} \logicalOR \T{LOWER} \\
\mbox{Position } k\quad &:= && -100 \logicalOR -99 \logicalOR \hdots \logicalOR 1 \logicalOR 2 \logicalOR \hdots \logicalOR 100 \\
\mbox{Index } i\quad &:= && -5 \logicalOR -4 \logicalOR \hdots \logicalOR -1 \logicalOR 1 \logicalOR 2 \logicalOR \hdots \logicalOR 5 \\
\mbox{Boundary } b\quad &:= && \T{START} \logicalOR \T{END} \\
\mbox{Delimiter } \delta\quad &:= && \&\,,.?@()[]\%\{\}/:;\$\#"' \\
\mbox{Character } c\quad &:= && A-Z \logicalOR a-z \logicalOR 0-9 \logicalOR \&,.?@\hdots
\end{alignat*}
    \caption{The DSL for string transformation tasks \citep{ROBUSTFILL}} 
    \label{fig:robustfill_dsl}
\end{figure}

The DSL for string transformations we use is the same as used in RobustFill \citep{ROBUSTFILL}, and is shown in \autoref{fig:robustfill_dsl}. The top-level operator for programs in the DSL is a $\T{Concat}$ operator that concatenates a random number (up to 10) of expressions $e_i$.
Each expression $e$ can either be a substring expression $f$, a nesting expression $n$, or a constant string $c$. A substring expression 
can either return the substring between left $k_1$ and right $k_2$ indices, or between the $i_1$-th occurence of regex $r_1$ and $i_2$-th occurence of regex $r_2$. The nesting expressions also return substrings of the input, such as extracting the $i$-th occurrence of a regex, but can also be composed with existing substring or nesting expressions for more complex string transformations.

\paragraph{RobustFill Model}  RobustFill \citep{ROBUSTFILL} is a seq-to-seq neural network that uses a encoder-decoder architecture where the encoder computes a representation of the input $e(X)$, and the decoder autoregressively generates the output given the source representation, i.e. conditional likelihood of $Y = [y_1, \hdots, y_T]$ decomposes as $p(Y|X) = \prod_{t=1}^T p(y_t|y_{<t}, X)$.

In RobustFill, the probability of decoding each token $y_t$ is given by $p(y_t |y_{<t}, X) = \softmax{W(h_t)}$
with $W$ being the projection onto logits, or unnormalized log probabilities.
The hidden representation $h_t$ is an LSTM hidden unit given by,
\begin{align*}
    E_t &= \attention{h_{t-1}, e(X)}, \\
    h_t &= \lstm{h_{t-1}, E_t}.
\end{align*}
Here $e(X)$ is the sequence of hidden states after processing the specifications with an LSTM encoder, and $\attention{Q, V}$ denotes the scaled dot-product attention with query $Q$ and key-value sequence $V$ \citep{Attention}. 
In the case of $X$ being multiple I/O examples, the RobustFill model of \citet{ROBUSTFILL} uses double attention
\begin{align*}
s_{t, i}^I &= \attention{h_{t-1}, e(I_i)} \\
s_{t, i}^O &= \attention{\concat{h_{t-1}, s_{t, i}^I}, e(O_i)} \\
h_{t, i} &=  \lstm{h_{t-1}, \concat{s_{t, i}^I, s_{t, i}^O}} \,\, \forall 1 \leq i \leq N,
\end{align*}
and hidden states are pooled across examples before being fed into the final softmax layer, or
$
h_t =  \mathrm{maxpool}_{1 \leq i \leq N} \tanh(V(h_{t, i}))
$
, where $V$ is another projection. 

\newpage
\section{Latent Programmer Architecture}
\label{sec:architecture_details}

Recall that the LP architecture consists of three modules: a program encoder, latent predictor, and latent program decoder.

\paragraph{Program Encoder} The program encoder $\aef(Y)$ is a Transformer encoder, followed by a stack of convolutions of stride $2$, each halving the size of the sequence. We apply the convolution $\ell$ times, which reduces a $T$-length program to a latent sequence of length $\lceil T / 2^\ell \rceil$. This provides temporal abstraction, since the high-level planning actions are made only every $2^\ell$ steps.
In summary, the program encoder is given by $\aef(Y) \leftarrow h_\ell$, where
\begin{align}
\begin{split}
    h_0 &\leftarrow \mathrm{TransformerEncoder}(Y) \\
    h_m &\leftarrow \mathrm{Conv}(h_{m-1}) \text{ for $m \in 1 \ldots \ell$} \\
\end{split}
\label{eq:lp_encoder}
\end{align}
Here $\mathrm{TransformerEncoder}(\cdot)$ applies a stack of self-attention and feed-forward units on input embeddings via a residual path, described in detail by~\citet{Transformer}.
This will be used, along with the latent program decoder, as an autoencoder during training (see~\autoref{sec:training}).

\paragraph{Latent Predictor} The latent predictor $\lp(X)$ is a Transformer that autoregressively outputs probabilities over latent tokens, which can be decoded using search algorithms such as beam search to generate a predicted latent code $Z'$. This is different than the program encoder, which outputs a single sequence $Z$, because we use the latent predictor to organize search over latent codes; at test time, we will obtain a $L$-best list of latent token sequences from $\lp(X)$.

\paragraph{Latent Program Decoder} The latent program decoder $d(Z, X)$ is a Transformer that jointly attends to the latent sequence and program specification, to autoregressively generates a distribution over program tokens.
This is performed via two separate attention modules, whose outputs are concatenated into the hidden unit. Formally, given a partially
generated program $Y' = [y'_1, y'_2, \hdots, y'_{t-1}]$, and the encoded specification $E = \mathrm{TransformerEncoder}(X)$, the latent program decoder performs
\begin{align}
\begin{split}
    e_t &\leftarrow \mathrm{TransformerDecoder}(Y', E)_{t-1} \\
    z_t &\leftarrow \mathrm{TransformerDecoder}(Y', Z)_{t-1} \\
    h_t &\leftarrow \concat{e_t, z_t},
\end{split}
\label{eq:lp_decoder}
\end{align}
where $\mathrm{TransformerDecoder}(x, y)$ denotes a Transformer decoder applied to outputs $y$ while attending to inputs encoding $x$, and the subscript indexes an entry in the resulting output sequence. Finally, the distribution over output token $k$ is given by $\softmax{W(h_t)},$ where $W$ is a learned
parameter matrix.
When $X$ is multiple I/O examples, each example is encoded as
$E_i = \mathrm{TransformerDecoder}(I_i, O_i)$. Then, a separate hidden state per I/O is computed following \eqref{eq:lp_decoder}, followed by a late max-pool to get the final hidden state.

\newpage
\section{Interpretability Experiments on Toy DSL}
\label{sec:interpretability}
\begin{figure}[ht]
\small
\begin{alignat*}{2}
\mbox{Program } Y\quad &:= &\quad& \T{Concat}(e_1, e_2, \hdots) \\
\mbox{Expression } e\quad &:= && \T{GetSpan}(r_1, i_1, r_2, i_2)  \\
\mbox{Regex } r\quad &:= && t_1 \logicalOR \hdots \logicalOR t_n \logicalOR \delta_1 \logicalOR \hdots \logicalOR \delta_m \\
\mbox{Type } t\quad &:= &&  \T{NUMBER} \logicalOR \T{WORD} \logicalOR \T{ALPHANUM} \\
\mbox{Index } i\quad &:= && -1 \logicalOR 1 \logicalOR 2 \\
\mbox{Delimiter } \delta\quad &:= && \&\,,. \\
\mbox{Character } c\quad &:= && A-Z \logicalOR a-z \logicalOR 0-9 \logicalOR \&\,,.
\end{alignat*}
    \caption{Toy DSL for string transformation tasks} 
    \label{fig:toy_dsl}
\end{figure}

\begin{figure}[H]
\begin{minipage}{\textwidth}
    \centering
    \begin{tabular}{p{5cm} p{8cm}}
     \toprule
     Inputs & Outputs \\
     \midrule
     ``,C,XoC" &  ``C" \\
     ``.G73,NT" & ``G73" \\
     ``.Uvg t7MXI" & ``Uvg" \\
     ``.tLqFJ .dMKlh" & ``tLqFJ" \\
    \bottomrule
    \end{tabular}
\end{minipage}
\begin{minipage}{\textwidth}
    \vspace{0.1in}
    \centering
    \begin{tabular}{p{1.5cm} | p{12.5cm}}
    \phantom{.}LP & \small\texttt{GetSpan\_ALPHANUM\_1\_ALPHANUM\_1} \\\rule{0pt}{3ex}

    LP Latent & \small\texttt{TOK\_6} \\
    \end{tabular}
\end{minipage}\vspace{0.4in}
\begin{minipage}{\textwidth}
    \centering
    \begin{tabular}{p{5cm} p{8cm}}
     \toprule
     Inputs & Outputs \\
     \midrule
     ``,3okM5,,," &  ``3okM5 ,3okM53okM5 ,3okM5" \\
     ``,, ,.O8p" & ``O8p ,.O8pO8p ,.O8p" \\
     ``,, , ,IBpU" & ``IBpU ,IBpUIBpU ,IBpU" \\
     ``,,,,mUV" & ``mUV ,,,,mUVmUV ,,,,mUV" \\
    \bottomrule
    \end{tabular}
\end{minipage}
\begin{minipage}{\textwidth}
    \vspace{0.1in}
    \centering
    \begin{tabular}{p{1.5cm} | p{12.5cm}}
    \phantom{.}LP & \small\texttt{GetSpan\_ALPHANUM\_1\_ALPHANUM\_1 | GetSpan\_,\_-1\_ALPHANUM\_1 | GetSpan\_ALPHANUM\_1\_ALPHANUM\_-1 | GetSpan\_,\_2\_ALPHANUM\_1} \\\rule{0pt}{3ex}
    LP Latent & \small\texttt{TOK\_6 | TOK\_5 | TOK\_6 | TOK\_5} \\
    \end{tabular}
\end{minipage}\vspace{0.4in}
\begin{minipage}{\textwidth}
    \centering
    \begin{tabular}{p{5cm} p{8cm}}
     \toprule
     Inputs & Outputs \\
     \midrule
     ``,CNBA,uJke.00 Hm 6938" &  ``CNBA,uJke.00CNBA,6938" \\
     ``.Xp.sYH ,46,Rj ,330" & ``Xp.sYH ,46Xp.sYH ,46,Rj ,330" \\
     ``,gYR 85296 LRgJX,15,eWEeu" & ``gYR 85296gYR 85296 LRgJX,15,15" \\
     ``.BPYVr ALVbf wEvm 86,103" & ``BPYVr ALVbf wEvm 86BPYVr ALVbf wEvm 86,103" \\
    \bottomrule
    \end{tabular}
\end{minipage}
\begin{minipage}{\textwidth}
    \vspace{0.1in}
    \centering
    \begin{tabular}{p{1.5cm} | p{12.5cm}}
    \phantom{.}LP & \small\texttt{GetSpan\_WORD\_1\_NUMBER\_1 | GetSpan\_WORD\_1\_,\_-1 | GetSpan\_NUMBER\_2\_NUMBER\_2} \\\rule{0pt}{3ex}
    LP Latent & \small\texttt{TOK\_9 | TOK\_9 | TOK\_4} \\
    \end{tabular}
\end{minipage}\vspace{0.4in}
\caption{Latent codes and programs found by Latent Programmer in toy DSL.}
\label{fig:toy_supp_examples}
\end{figure}

\begin{figure}[H]
\begin{minipage}{\textwidth}
    \centering
    \begin{tabular}{p{5cm} p{8cm}}
     \toprule
     Inputs & Outputs \\
     \midrule
     ``r, 6150,XLQPl" &  ``6150r, 6150r, 6150" \\
     ``.ERYlM, 80,Iejg" & ``80ERYlM, 80ERYlM, 80" \\
     ``sqd,.xJx,01928" & ``01928sqd,.xJx,01928sqd,.xJx,01928" \\
     ``.w Nqk.42," & ``42w Nqk.42w Nqk.42" \\
    \bottomrule
    \end{tabular}
\end{minipage}
\begin{minipage}{\textwidth}
    \vspace{0.1in}
    \centering
    \begin{tabular}{p{1.5cm} | p{12.5cm}}
    \phantom{.}LP & \small\texttt{GetSpan\_NUMBER\_-1\_NUMBER\_-1 | GetSpan\_WORD\_1\_NUMBER\_1 | GetSpan\_WORD\_1\_,\_-1} \\\rule{0pt}{3ex}
    LP Latent & \small\texttt{TOK\_4 | TOK\_9 | TOK\_9} \\
    \end{tabular}
\end{minipage}\vspace{0.4in}
\begin{minipage}{\textwidth}
    \centering
    \begin{tabular}{p{5cm} p{8cm}}
     \toprule
     Inputs & Outputs \\
     \midrule
     ``.VyPL 3785.0933,Xj EFSjp" &  ``VyPL37853785.0933,Xj EFSjp" \\
     ``.023 Jz Suz.t .4" & ``Jz023023 Jz Suz.t" \\
     ``TyCBs,803 TjtA,4 .qH" & ``TyCBs803803 TjtA,4 .qH" \\
     ``.cCr,3248 L ,QPLd.6472" & ``cCr32483248 L ,QPLd" \\
    \bottomrule
    \end{tabular}
\end{minipage}
\begin{minipage}{\textwidth}
    \vspace{0.1in}
    \centering
    \begin{tabular}{p{1.5cm} | p{12.5cm}}
    \phantom{.}LP & \small\texttt{GetSpan\_WORD\_1\_WORD\_1 | GetSpan\_NUMBER\_1\_NUMBER\_1 | GetSpan\_NUMBER\_1\_WORD\_-1} \\\rule{0pt}{3ex}
    LP Latent & \small\texttt{TOK\_6 | TOK\_7 | TOK\_9} \\
    \end{tabular}
\end{minipage}\vspace{0.4in}
\begin{minipage}{\textwidth}
    \centering
    \begin{tabular}{p{5cm} p{8cm}}
     \toprule
     Inputs & Outputs \\
     \midrule
     ``.Eu.F IgKFs,XD.011" &  ``011F" \\
     ``.U0Z,aVEzk,KNq 08,UqlhR" & ``0Z" \\
     ``44 j.Oz.peQy,l" & ``44Oz" \\
     ``,FtAz CIHLB V 851.oR8l" & ``851CIHLB" \\
    \bottomrule
    \end{tabular}
\end{minipage}
\begin{minipage}{\textwidth}
    \vspace{0.1in}
    \centering
    \begin{tabular}{p{1.5cm} | p{12.5cm}}
    \phantom{.}LP & \small\texttt{GetSpan\_NUMBER\_1\_NUMBER\_1 | GetSpan\_WORD\_2\_WORD\_2} \\\rule{0pt}{3ex}
    LP Latent & \small\texttt{TOK\_7 | TOK\_3} \\
    \end{tabular}
\end{minipage}\vspace{0.4in}
\begin{minipage}{\textwidth}
    \centering
    \begin{tabular}{p{5cm} p{8cm}}
     \toprule
     Inputs & Outputs \\
     \midrule
     ``.9312 ..767" &  ``7679312 ..76793129312" \\
     ``.,04194,47460" & ``4746004194,474600419404194" \\
     ``.4940..3646" & ``36464940..364649404940" \\
     ``. .180,5275" & ``5275180,5275180180" \\
    \bottomrule
    \end{tabular}
\end{minipage}
\begin{minipage}{\textwidth}
    \vspace{0.1in}
    \centering
    \begin{tabular}{p{1.5cm} | p{12.5cm}}
    \phantom{.}LP & \small\texttt{GetSpan\_NUMBER\_2\_NUMBER\_-1 | GetSpan\_NUMBER\_1\_NUMBER\_-1 | GetSpan\_NUMBER\_1\_NUMBER\_1 | GetSpan\_NUMBER\_1\_NUMBER\_1} \\\rule{0pt}{3ex}
    LP Latent & \small\texttt{TOK\_4 | TOK\_9 | TOK\_7 | TOK\_7} \\
    \end{tabular}
\end{minipage}\vspace{0.4in}
\caption{More latent codes and programs found by Latent Programmer in toy DSL.}
\label{fig:toy_supp_examples_2}
\end{figure}

\newpage\newpage
\section{Examples of Generated Programs and Latent Codes}
\label{sec:supp_examples}
\begin{figure}[H]
\begin{minipage}{\textwidth}
    \centering
    \begin{tabular}{p{6cm} p{3.5cm} p{3.5cm}}
     \toprule
     Inputs & Outputs & LP Outputs \\
     \midrule
     ``Mason Smith" &  ``Smith M" & ``Smith M"\\
     ``Henry Myers" & ``Myers H" & ``Myers H" \\
     ``Barry Underwood" & ``Underwood B" & ``Underwood B" \\
     ``Sandy Jones" & ``Jones S" & ``Jones S"\\
    \bottomrule
    \end{tabular}
\end{minipage}
\begin{minipage}{\textwidth}
    \vspace{0.1in}
    \centering
    \begin{tabular}{p{1.5cm} | p{12.5cm}}
    \phantom{.}LP & \small\texttt{GetToken\_PROP\_CASE\_2 | ConstStr(`` ") | GetToken\_CHAR\_1(GetToken\_PROP\_CASE\_1)} \\\rule{0pt}{3ex}
    LP Latent & \small\texttt{TOK\_30 | TOK\_13 | TOK\_39 | TOK\_30} \\
    \end{tabular}
\end{minipage}\vspace{0.4in}
\begin{minipage}{\textwidth}
    \centering
    \begin{tabular}{p{6cm} p{3.5cm} p{3.5cm}}
     \toprule
     Inputs & Outputs & LP Outputs \\
     \midrule
     ``January 15" &  ``jan 15" & ``jan 15" \\
     ``febuary 28" & ``feb 28" & ``feb 28" \\
     ``march 1" & ``mar 1" & ``mar 1" \\
     ``October 31" & ``oct 31" & ``oct 31" \\
    \bottomrule
    \end{tabular}
\end{minipage}
\begin{minipage}{\textwidth}
    \vspace{0.1in}
    \centering
    \begin{tabular}{p{1.5cm} | p{12.5cm}}
    \phantom{.}LP & \small\texttt{ToCase\_LOWER(SubStr(1, 3)) | ConstStr(`` ") | GetToken\_NUMBER\_1} \\\rule{0pt}{3ex}
    LP Latent & \small\texttt{TOK\_11 | TOK\_26 | TOK\_17} \\
    \end{tabular}
\end{minipage}\vspace{0.4in}
\begin{minipage}{\textwidth}
    \centering
    \begin{tabular}{p{6cm} p{3.5cm} p{3.5cm}}
     \toprule
     Inputs & Outputs & LP Outputs \\
     \midrule
     ``(321) 704 3331" &  ``321.704.3331" & ``321.704.3331"\\
     ``(499) 123 3574" & ``499.123.3574" & ``499.123.3574" \\
     ``(555) 580 8390" & ``555.580.8390" & ``555.580.8390" \\
     ``(288)225 6116" & ``288.225.6116" & ``288.225.6116"\\
    \bottomrule
    \end{tabular}
\end{minipage}
\begin{minipage}{\textwidth}
    \vspace{0.1in}
    \centering
    \begin{tabular}{p{1.5cm} | p{12.5cm}}
    \phantom{.}LP & \small\texttt{GetToken\_NUMBER\_1 | ConstStr(.) | Replace\_`` "\_.(SubStr(-8, -1))} \\\rule{0pt}{3ex}
    LP Latent & \small\texttt{TOK\_17 | TOK\_27 | TOK\_24 | TOK\_16} \\
    \end{tabular}
\end{minipage}\vspace{0.4in}
\begin{minipage}{\textwidth}
    \centering
    \begin{tabular}{p{6cm} p{3.5cm} p{3.5cm}}
     \toprule
     Inputs & Outputs & LP Outputs \\
     \midrule
     ``Milk 4, Yoghurt 12, Juice 2, Egg 5" &  ``M.E." & ``M.E."\\
     ``US:38 China:35 Russia:27 India:1" & ``U.I." & ``U.I." \\
     ``10 Apple 2 Oranges 13 Bananas 40 Pears" & ``A.P." & ``A.P." \\
     ``parul 7 rico 12 wolfram 15 rick 19" & ``P.R." & \textcolor{red}{``.."}\\
    \bottomrule
    \end{tabular}
\end{minipage}
\begin{minipage}{\textwidth}
    \vspace{0.1in}
    \centering
    \begin{tabular}{p{1.5cm} | p{12.5cm}}
    \phantom{.}LP & \small\texttt{GetToken\_CHAR\_1(GetToken\_PROP\_CASE\_1) | Const(.) | GetToken\_CHAR\_-1(GetAll\_ALL\_CAPS) | Const(.)} \\\rule{0pt}{3ex}
    LP Latent & \small\texttt{TOK\_39 | TOK\_30 | TOK\_6 | TOK\_38 | TOK\_30} \\
    \end{tabular}
\end{minipage}
\caption{Latent codes and programs found by Latent Programmer in string transformation tasks. Red denotes I/O where the predicted program mapped input to an incorrect output.}
\label{fig:robustfill_supp_examples}
\end{figure}

\begin{figure}[H]
\centering
\begin{tabular}{p{5cm} p{8.5cm}} 
 \toprule
 Docstring & Program\\[0.1cm]
 \midrule
 get an environment variable &  \small\texttt{def \textcolor{red}{set\_key}(key, \textcolor{red}{val}, key\_prefix=None):} \\
 & \quad\small\texttt{return environ.get(key, key\_prefix)}\\
 \midrule
 return a list of the words & \small\texttt{def split(s, sep=None, maxsplit=-1):} \\
 in the string s & \quad\small\texttt{return s.split(sep, maxsplit)} \\
 \midrule
 mean squared error function &  \small\texttt{def mean\_squared\_error(y\_true, y\_pred):} \\
 & \quad\small\texttt{return tf.reduce\_mean(tf.square((y\_true - y\_pred)))}\\
 \midrule
 read a python file & \small\texttt{def read\_file(fname):} \\
 & \quad\small\texttt{\textcolor{red}{f = open(fname)}} \\
 & \quad\small\texttt{with open(fname,  `r') as f:} \\
 & \quad\quad\small\texttt{\textcolor{red}{f.seek(0)}} \\
 & \quad\quad\small\texttt{return f.read()} \\
 \midrule
 pickle dump & \small\texttt{def pickle\_\textcolor{red}{save}(filename, data):} \\
 & \quad\small\texttt{with open(filename, \textcolor{red}{`r'}) as f:} \\
 & \quad\quad\small\texttt{pickle.dump(data, f)} \\
 \midrule
 takes a timedelta and returns the& \small\texttt{def total\_seconds(delta):} \\
 total number of seconds & \quad\small\texttt{return ((delta.\textcolor{red}{microseconds} + ((delta.days * 24) * 3600) \textcolor{red}{* (10**6))/(10**6))}} \\
\bottomrule
\end{tabular}
\caption{Programs found by Latent Programmer in Python code generation dataset. Red denotes ares where the predicted program deviates from human code.}
\label{fig:python_supp_examples}
\vspace{-0.1in}
\end{figure}

\end{document}